\documentclass[10pt,twocolumn,letterpaper]{article}

\usepackage[pagenumbers]{cvpr} %

\pdfoutput=1
\usepackage[utf8]{inputenc} 
\usepackage[english]{babel} 
\usepackage[protrusion=true,expansion=true]{microtype}

\usepackage{overpic}
\usepackage{color}
\usepackage[nice]{nicefrac}
\usepackage{algpseudocode}
\usepackage[linesnumbered,ruled,vlined]{algorithm2e}

\SetCommentSty{mycommfont}
\usepackage{algpseudocode}
\usepackage{multirow}
\usepackage{adjustbox}
\usepackage[font=small,labelfont=bf]{caption}
\usepackage{subcaption}
\usepackage{cancel}
\usepackage{currfile}
\usepackage{diagbox}

\usepackage{wrapfig}

\definecolor{turquoise}{cmyk}{0.65,0,0.1,0.3}
\definecolor{purple}{rgb}{0.65,0,0.65}
\definecolor{dark_green}{rgb}{0, 0.7, 0}
\definecolor{orange}{rgb}{0.8, 0.6, 0.2}
\definecolor{red}{rgb}{0.8, 0.2, 0.2}
\definecolor{darkred}{rgb}{0.6, 0.1, 0.05}
\definecolor{blueish}{rgb}{0.0, 0.3, .6}
\definecolor{light_gray}{rgb}{0.7, 0.7, .7}
\definecolor{pink}{rgb}{1, 0, 1}
\definecolor{greyblue}{rgb}{0.25, 0.25, 1}

\newcommand{\silvianew}[1]{{\color{black}{#1}}} %

\newcommand{\hidden}[1]{#1}

\newcommand{\Figure}[1]{Figure~\ref{fig:#1}}

\newcommand{\eq}[1]{(\ref{eq:#1})}

\newcommand{\Section}[1]{Section~\ref{sec:#1}}

\usepackage{blindtext}
\usepackage{bbm}

\usepackage{lipsum}

\renewcommand{\paragraph}[1]{\vspace{.5em}\noindent\textit{#1} --}

\newcommand{\algoName}{BayesRays\xspace}

\renewcommand{\paragraph}[1]{\vspace{.5em}\noindent\textbf{#1}.}

\DeclareMathOperator*{\argmin}{arg\,min}
\DeclareMathOperator*{\argmax}{arg\,max}

\newcommand{\expect}{\mathbb{E}}

\newcommand{\given}{|}

\newcommand{\nRays}{R}
\newcommand{\radiance}{\mathbf{c}}
\newcommand{\rfield}{\mathcal{R}}
\newcommand{\density}{\tau}
\newcommand{\x}{\mathbf{x}}
\newcommand{\dir}{\mathbf{d}}

\newcommand{\nerfparams}{\boldsymbol{\phi}}
\newcommand{\nerfparamsOptimal}{{\nerfparams^{*}}}
\newcommand{\params}{\boldsymbol{\theta}}
\newcommand{\ray}{\mathbf{r}}

\newcommand{\raycolor}{\mathbf{C}}
\newcommand{\image}{\mathbf{I}}
\newcommand{\gt}{\text{gt}}
\newcommand{\datasetsize}{\mathbf{N}}

\newcommand{\logposterior}{h}

\newcommand{\Hessian}{\mathbf{H}}

\newcommand{\Fisher}{\mathcal{I}}
\newcommand{\expected}{\mathbb{E}}

\newcommand{\ufield}{\mathcal{U}}

\newcommand{\cameraparams}{\{\mathbf{T}_n\}}

\newcommand{\deformfield}{\mathcal{D}}
\newcommand{\deformparam}{\boldsymbol{\theta}}
\newcommand{\deformparamOptimal}{\theta^*}
\newcommand{\deformparamset}{\Theta}

\newcommand{\hessian}{\mathbf{H}}
\newcommand{\identity}{\mathbf{I}}

\newcommand{\residual}{\boldsymbol{\epsilon}}
\newcommand{\J}{\mathbf{J}}

\newcommand{\gtcolor}{\mathbf{y}}
\newcommand{\covariance}{\boldsymbol{\Sigma}}

\newcommand{\gridSize}{M}
\newcommand{\images}{\mathcal{I}}

\definecolor{cvprblue}{rgb}{0.21,0.49,0.74}
\usepackage[pagebackref,breaklinks,colorlinks,citecolor=cvprblue]{hyperref}

\def\paperID{*****} %
\def\confName{CVPR}
\def\confYear{2024}

\title{\LARGE \bf
Bayes' Rays: Uncertainty Quantification for Neural Radiance Fields}

\author{Lily Goli$^{1}$ \:\: Cody Reading$^{2}$ \:\: Silvia Sellán$^{1}$ \:\: Alec Jacobson$^{1,4}$ \:\: Andrea Tagliasacchi$^{1,2,3}$ \\[.5em] 
University of Toronto$^{1}$ \:\: Simon Fraser University$^{2}$ \:\: Google DeepMind$^{3}$ \: Adobe Research$^{4}$
}

\begin{document}

\twocolumn[{
\renewcommand\twocolumn[1][]{#1}%
\maketitle
\begin{center}
\centering
\captionsetup{type=figure}
\vspace{-0.5cm}
\begin{overpic} 
    [width=.99\linewidth]
    {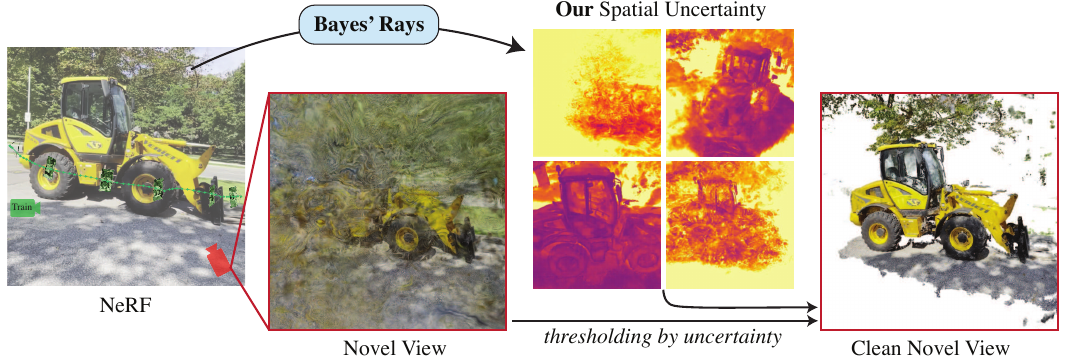}

\end{overpic}
\vspace{-0.4cm}
\captionof{figure}{
We introduce \algoName, a post-hoc algorithm to estimate the spatial uncertainty of any pre-trained NeRF of any arbitrary architecture. Our method requires no additional training and can be used to clean up NeRF artifacts caused by occluded or incomplete data.
}
\label{fig:teaser}
\vspace{-0.3cm}
\end{center}
}]

\begin{abstract}
Neural Radiance Fields (NeRFs) have shown promise in applications like view synthesis and depth estimation, but learning from multiview images faces inherent uncertainties. Current methods to quantify them are either heuristic or computationally demanding. We introduce $\algoName$, a post-hoc framework to evaluate uncertainty in any pre-trained NeRF without modifying the training process. Our method establishes a volumetric uncertainty field using spatial perturbations and a Bayesian Laplace approximation. We derive our algorithm statistically and show its superior performance in key metrics and applications. Additional results available at: \url{https://bayesrays.github.io}
\end{abstract}

\vspace{-0.4cm}
\section{Introduction}
\label{sec:intro}

Neural Radiance Fields (NeRFs) are a class of learned volumetric implicit scene representations that have exploded in popularity due to their success in applications like novel view synthesis and depth estimation.
The process of learning a NeRF from a discrete set of multiview images is plagued with uncertainty: even in perfect experimental conditions, occlusions and missing views will limit the epistemic knowledge that the model can acquire about the scene. 

\silvianew{Studying the epistemic uncertainty in NeRF is fundamental for tasks like outlier detection~\cite{marimont2021implicit} and next-best-view planning~\cite{pan2022activenerf} that expand NeRF's performance and application domain to critical areas like autonomous driving~\cite{chitta2021neat}.}
However, quantifying the uncertainty contained in a NeRF model is a relatively new area of study, with existing methods proposing either heuristic proxies without theoretical guarantees \hidden{\cite{zhan2022activermap,probnerf}} or probabilistic techniques that require costly computational power \hidden{\cite{sunderhauf2022densityaware}} and/or elaborate changes to the conventional NeRF training pipeline \hidden{\cite{stochasticnerf,cfnerf}}.

\begin{figure}[b]
    \includegraphics[width=\columnwidth]{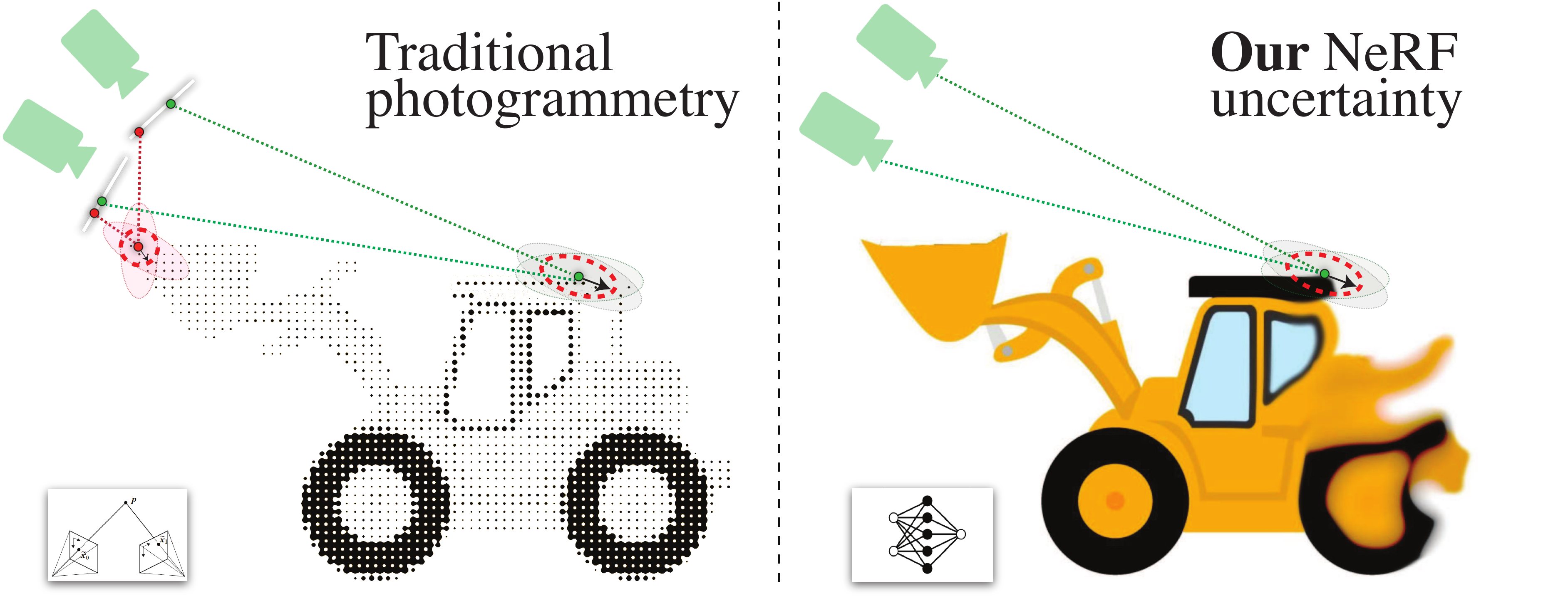}
    \vspace{-0.6cm}
    \caption{Inspired by uncertainty quantification in classic photogrammetry (left), we find epistemic uncertainty in NeRF (right).}
    \label{fig:idea}
\end{figure}

We draw inspiration from triangulation problems in classic photogrammetry \cite{szeliski2022computer}, where uncertainty is often modeled through distributions of feature point positions in image space that are then projected onto 3D (see \Figure{idea}). Intuitively, this distribution measures how much a feature's position can be perturbed while maintaining multi-view consistency. We apply a similar intuition to NeRF, identifying the regions of the radiance field that can be spatially perturbed with minimal impact on the reconstruction loss.

\begin{figure}
    \includegraphics[width=\linewidth]{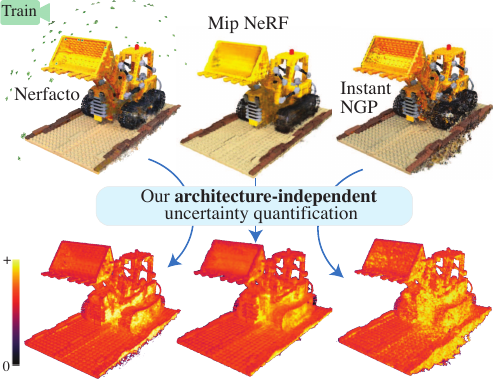}
    \vspace{-0.8cm}
    \caption{\algoName applied to different NeRF architectures. "Lego" \cite{mildenhall2020nerf} is only trained with cameras on its left hemisphere.}
    \label{fig:ablation1}
\end{figure}

We propose \algoName, a post-hoc framework for quantifying the uncertainty of any arbitrary pre-trained NeRF. Without requiring any changes to the training pipeline and regardless of the architecture in use in the given NeRF (see \Figure{ablation1}), our method simulates spatially parametrized perturbations of the radiance field and uses a Bayesian Laplace approximation to produce a volumetric \emph{uncertainty field}, which can be rendered like an additional color channel.

In this work, we show that our calculated uncertainties are not only statistically meaningful but also outperform previous works on key metrics like correlation to reconstructed depth error. Furthermore, they provide a framework for critical applications like removing \textit{floater} artifacts from NeRF, matching or improving the state-of-the-art (see~\Figure{teaser}).

\noindent In summary, our main contributions are:
\begin{itemize}
    \item We introduce a plug-and-play probabilistic method to quantify the uncertainty of any pre-trained Neural Radiance Field independently of its architecture, and without needing training images or costly extra training. 
    \item In little over a minute, we compute a spatial uncertainty field that can be rendered as an additional color channel.
    \item We propose thresholding our uncertainty field to remove artifacts from pre-trained NeRFs interactively in real time.
\end{itemize}

\section{Related work}
\label{sec:related}
Uncertainty Quantification studies the distribution of the responses of a system conditioned on a set of measurable input variables \cite{smith2013uncertainty}. As a field of statistics, it has grown over many decades out of the need to measure the accuracy of scientific predictions in areas like physics \cite{ellis1980uncertainties}, chemistry \cite{schecher1988evaluation} or meteorology \cite{palmer2000predicting}.

\paragraph{Uncertainty in Computer Vision} Closer to our application, estimating the uncertainty of Computer Vision systems has been a subject of study even long before the Deep Learning revolution; for example, in Structure from Motion and Bundle Adjustment \cite[Section 11.4]{szeliski2022computer}\cite{Triggs1999BundleA}. In these problems from classic photogrammetry, scene geometry and camera parameters are jointly optimized in a process filled with uncertainty \cite{Szeliski1997, Wilson2020VisualizingSB}, often modeled through 2D image-space Gaussians projected to 3D \cite{szeliski2022computer, Morris1999UncertaintyMF,Triggs1999BundleA, Bartoli2003} (see \Figure{idea}).

\paragraph{Uncertainty in Deep Learning} The process of fitting a neural network to observed data typically contains two fundamentally different sources of uncertainty \cite{uncertaintyReview2021,kendall2017uncertainties}. 
\emph{Aleatoric} uncertainty refers to the inherent randomness contained in the data (e.g., due to instrument error or uncontrolled influences) and is often captured by fitting not just one function but a parametric distribution (e.g., Gaussians) to the data~\cite{monteiro2020stochastic,ayhan2022test,kendall2017uncertainties}. On the other hand,
\emph{epistemic} uncertainty quantifies the lack of knowledge that a model has over the system it is trying to replicate; for example, due to missing data. Commonly, this is achieved through a Bayesian framework that estimates the posterior distribution of the model given observed data. The most straightforward-yet-expensive way of achieving this is via the use of \emph{Deep Ensembles} that quantify the differences in optimized parameters after training many identical, yet differently initialized, networks on the same data \cite{ensemble2016, ensemble2021}. A popular alternative to ensembles are variational Bayesian Neural Networks, which model each network parameter with an independent distribution that is sampled at each training iteration and adjusted through a KL loss to approximate the true model posterior \cite{Neal1995BayesianLF, blundell2015weight}.
This significant change to the training pipeline comes at a computational cost, which recent works have proposed circumventing through the use of post-hoc \emph{Laplace Approximations} \silvianew{(see \cite{Ritter2018ASL, Kristiadi2021}, or \cite{tkach2017online} for a Computer Vision application)} that only estimate the network weight posterior near their already-trained value using derivative information. 

\paragraph{Uncertainty in Neural Radiance Fields}
NeRFs~\cite{NeRF} represent 3D scenes through a neural volumetric encoding that is optimized to match the images produced from multiple camera views. Aleatoric uncertainty presents itself in this process through the presence of transient objects in the scene or changes in lighting and camera specifications. These phenomena are quantified by the pioneering work NeRF-W  \cite{nerfwild} and subsequent follow-ups \cite{pan2022activenerf, Jin2023neunbv, Ran_2023} through a combination of standard aleatoric Deep Learning techniques \cite{uncertaintyReview2021} and a learned appearance latent embedding~\cite{bojanowski2019optimizing}.

Distinctly, we concern ourselves with the epistemic uncertainty of Neural Radiance Fields, the source of which is often missing data due to occlusions, ambiguities and limited camera views. Many of the general Deep Learning techniques to quantify this uncertainty have been applied to NeRF with limited success. Works like \cite{sunderhauf2022densityaware} propose uncertainty estimation through ensemble learning, which can be time and memory consuming. 
\citet{stochasticnerf} and its follow-up~\cite{cfnerf} model the problem through variational inference and KL divergence optimization in a way that is not too dissimilar in principle, yet shown to be superior, to standard variational Bayesian neural networks.
All these methods require intricate changes to the NeRF training pipeline. In contrast, we introduce \algoName , the first framework that allows the use of Laplace approximations for NeRF uncertainty quantification, avoids variational optimization and can thus be applied on any pretrained NeRF of any arbitrary pipeline.

Away from the traditional Deep Learning uncertainty quantification frameworks, other works propose using NeRF-specific proxies for uncertainty. For example, \citet{zhan2022activermap} propose computing the uncertainty as entropy of ray termination in NeRF model. While high entropy can be a good indicator of uncertainty in modeling solid objects, such assumption can fail while using density regularizers like the distortion loss proposed in \cite{mipnerf-360}.
\citet{probnerf} suggest quantifying uncertainty as the variance in scenes produced by a generative model when conditioned on partial observations, relying heavily on the specific model's priors.

\paragraph{Other related works} As we show in Section~\ref{sec:method}, our uncertainty quantification relies on the sensitivity of any trained NeRF to perturbations, a concept recently explored \silvianew{in an unpublished preprint by \citet{yan2023active}} for applications outside NeRF and through a continual learning framework. To make our method computationally tractable and architecture-independent, we introduce a spatial deformation field similar to the one suggested by \cite{nerfies, park2021hypernerf}, albeit we interpret it instead as a re-parametrization of the NeRF model on which to perform a Laplace Approximation to quantify its uncertainty. One of the many uses we propose for our output spatial uncertainty field is to avoid common spatial NeRF artifacts. Instead of changing the optimization at training time by adding regularizers~\cite{niemeyer2021regnerf, mipnerf-360, park2023camp, rebain2022lolnerf}, we propose removing them from any pre-trained NeRF in a post-processing step, a task that has been tackled recently by diffusion-based works like Nerfbusters~\cite{warburg2023nerfbusters}.
As we show in Section~\ref{sec:results}, our algorithm matches or improves the performance of Nerfbusters in this specific application while being more general and requiring no additional training. 

\begin{figure}[t]
\includegraphics[width=\columnwidth]{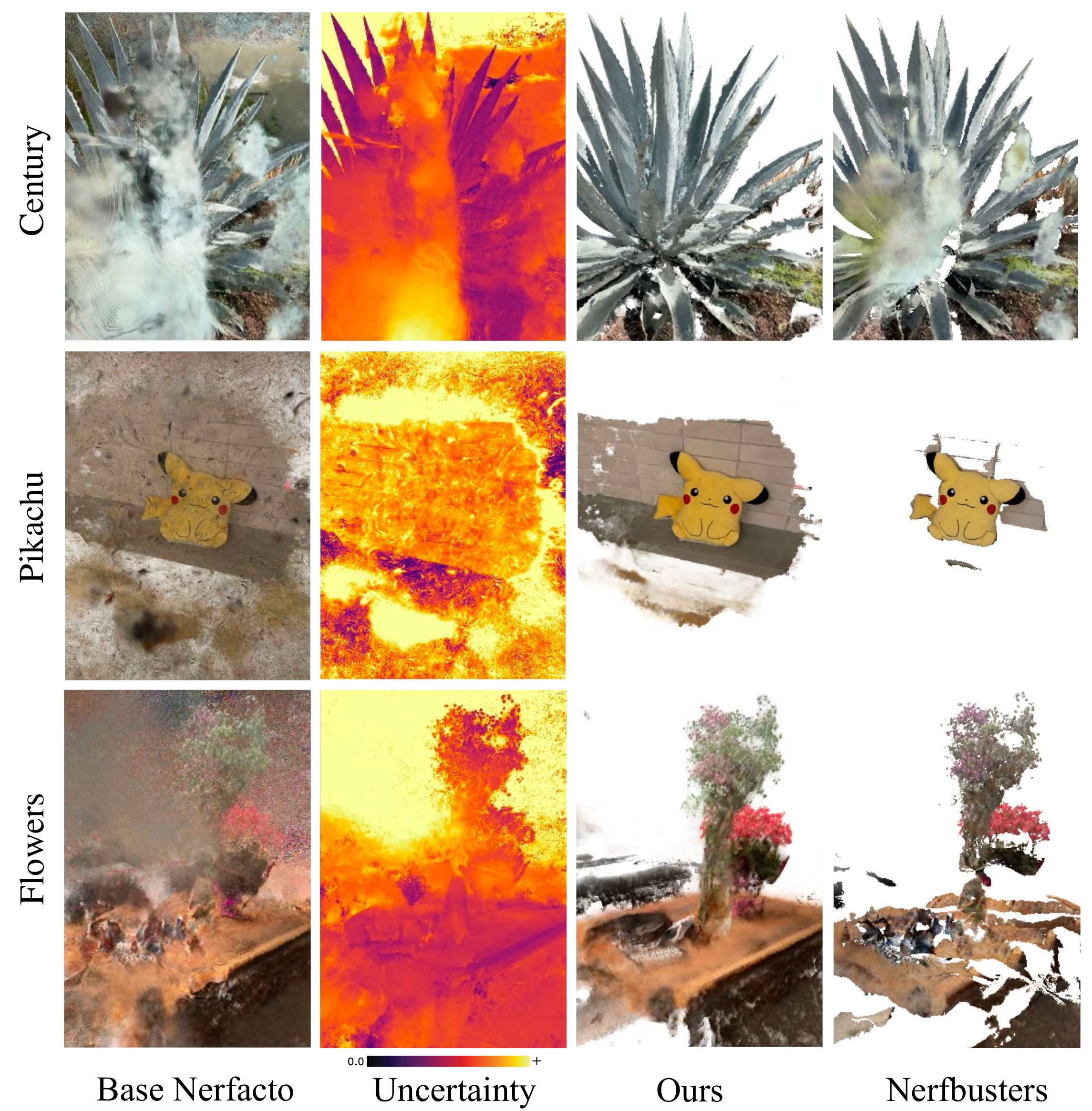}
\\[0.7em]
\centering
\resizebox{\linewidth}{!}{
\begin{tabular}{c|c|c|c|c}
        Method & PSNR $\uparrow$ & SSIM $\uparrow$ & LPIPS  $\downarrow$ & Coverage $\uparrow$ \\ \hline \hline
        Nerfacto (base) & 16.83 & 0.52 & 0.39 & 0.89 \\ \hline  \hline
        Nerfbusters & 17.99 &  \textbf{0.60} &  \textbf{0.25} & 0.63 \\ \hline
        $\algoName$-0.9 & 17.66 & 0.56 & 0.34 &  \textbf{0.83} \\ \hline
        $\algoName$-0.4 & 17.78 & 0.57 & 0.31 & 0.78 \\ \hline
        $\algoName$-best & \textbf{18.27} &  \textbf{0.60} & 0.27 & 0.70 \\ \hline
\end{tabular}
}

\vspace{-0.2cm}
\caption{
We propose cleaning up a learned scene by thresholding it based on our computed uncertainty, matching or surpassing the state of the art at a much lower computational and memory cost.} %
\label{fig:nerfbuster-comparison}
\end{figure}

\section{Background}
\label{sec:background}
We propose a general framework for applying Laplace approximations to quantify the epistemic uncertainty of any pre-trained NeRF. We will briefly review both these concepts before exploring the difficulties one encounters when trying to combine them naively, thus motivating our perturbation-based approach described in Sec.~\ref{sec:method}.

\subsection{Neural Radiance Fields}
\label{sec:nerf}
Conventional NeRFs \cite{NeRF} learn to map each point in 3D space to a view-dependent radiance and a view-independent density value:
\begin{equation}
\radiance_{\nerfparams}(\x, \dir) ,\:
\density_{\nerfparams}(\x)
= \rfield(\x, \dir ; \nerfparams)
\end{equation}
where $\nerfparams$ represent the learnable parameters in the neural field. The color of each pixel in an image can then be rendered through compositing the density and color of a series of points $\{t_i\}$ along the ray $\ray = \mathbf{o}_\ray + t \cdot \mathbf{d}_\ray $, using volume rendering~\cite{digest}:
\begin{equation}
     \raycolor_{\nerfparams}(\ray) = \sum_i \exp\left(-\sum_{j<i} \tau_j \delta_j\right)(1-\exp(-\tau_i\delta_i)) \: \radiance_i,
\end{equation}
where $\delta_i$ denotes the distance between each pair of successive points. The network parameters $\nerfparams$ are optimized to minimize reconstruction loss defined as the squared distance between the predicted color $\raycolor(\ray)$ and ground truth $\raycolor^\gt$ for each ray $\ray$ sampled from image $\image_n$ of training set images $\images = \{\image\}_{n=0}^{\datasetsize}$. From a Bayesian perspective, this is equivalent to assuming a Gaussian likelihood $p(\raycolor_{\nerfparams}\given\nerfparams)\sim \mathcal{N}(\raycolor^\gt_n,\tfrac{1}{2})$ and finding $\nerfparamsOptimal$, the mode of the posterior distribution
\begin{equation}
    \nerfparamsOptimal = \argmax_{\nerfparams} p(\nerfparams\given \images) 
\end{equation}
which, by Bayes' rule, is the same as minimizing the negative log-likelihood
\begin{equation}
\nerfparamsOptimal = \argmin_{\nerfparams}
\:\:
\expect_{i} \,
\expect_{\ray \sim \image_n}
\| \raycolor_{\nerfparams}(\ray) - \raycolor^\gt_n(\ray) \|_2^2 
\label{eq:nerfrgb}
\end{equation}

\subsection{Neural Laplace Approximations}
\label{sec:lap}
A common strategy to quantify the epistemic uncertainty of any neural network trained on some data $\images$ is to study the posterior distribution of the network parameters $\params$ conditioned on the data, $p(\params \given \images)$. 
In contrast to variational Bayesian neural networks, which propose using Bayes' rule and a variational optimization to estimate this distribution, Laplace approximations \cite{Ritter2018ASL, Kristiadi2021} rely on simply training the network by any conventional means until convergence; i.e., on obtaining the likeliest network weights $\params^*$ -- the mode of $p(\params\given I)$.
Then, the posterior is approximated by a multivariate Gaussian distribution centered at the obtained mode $p(\params\given I) \sim \mathcal{N}(\params^*,\Sigma)$.
The covariance $\covariance$ of this distribution is then computed via a second-order Taylor expansion of the negative log-likelihood $\logposterior(\params) = - \log p(\params\given I)$ about $\params^*$:
\begin{equation}
\logposterior(\params) \approx \logposterior(\params^*) + \frac{1}{2}(\params{-}\params^*)^\top \Hessian(\params^*) \;(\params{-}\params^*)\,,
\label{eq:laplace}
\end{equation}
where first order terms are discarded since $\params^*$ is a maximum of $\logposterior(\params)$ and $\Hessian(\params^*)$ is the Hessian matrix of second derivatives of $\logposterior(\params)$ evaluated at $\params^*$.
Identifying the terms in Eq.~\ref{eq:laplace} with the usual log squared exponential Gaussian likelihood of $\mathcal{N}(\params^*,\covariance)$, one obtains
\begin{equation}
    \covariance = - \Hessian(\params^*)^{-1}
\end{equation}

Unfortunately, a naive application of this framework to NeRF by identifying $\params$ with $\nerfparams$ is impracticable, as it would have three potentially fatal flaws:
\begin{itemize}
\item First, as we show in Section~\ref{subsec:spatial-uncertainty}, the parameters of the NeRF are strongly correlated with each other, making it difficult to accurately estimate the posterior distribution with any guarantees without computing, (and storing) all the entries in $\Hessian$, a (potentially fully, at least block-) dense matrix with dimensions matching the number of network weights, before carrying out a costly inversion step. 
\item Secondly, even if one perfectly computed $\covariance$, the parameter correlations and network non-linearities would make it such that transferring this distribution to any geometrically meaningful one like a distribution over densities or pixel values would require repeatedly and expensively drawing samples from the full~$\mathcal{N}(\nerfparams^*,\covariance)$. 
\item Finally, beyond computational constraints, estimating uncertainty directly on the NeRF parameters would require our algorithm to have knowledge of (and, potentially, dependence on) the specific internal NeRF architecture used.
\end{itemize}

Below, we solve all of these problems by introducing a parametric \emph{perturbation field} on which to perform the Laplace approximation. Our algorithm is completely independent of the specific NeRF architecture used and can guarantee minimal correlations between parameters, allowing us to calculate a meaningful spatial uncertainty field without the need to draw any distribution samples.

\section{Method}
\label{sec:method}
As input, we assume that are given a pre-trained radiance field $\rfield$ with radiance function $\radiance$, density $\density$ and optimized parameters~$\nerfparamsOptimal$, as well as ground truth camera parameters~$\cameraparams$ corresponding to the~$\datasetsize$ training images.
Our method makes no assumption about the specific architecture of~$\rfield$ and is designed for any arbitrary framework that produces a learned density $\density_\nerfparamsOptimal$ and radiance $\radiance_\nerfparamsOptimal$, which we will treat as differentiable black boxes.

We begin by noting that, while the neural network weights~$\nerfparams$ may serve as a useful parametrization of~$\rfield$ during training, a Laplace approximation can be formulated on any re-parametrization~$\rfield_{\deformparam}$ for any parameter set~$\deformparam\in\deformparamset$, as long as one knows the mode of the distribution $p(\deformparam \given \images)$.

We follow by taking inspiration from the key insight behind NeRFs themselves: namely, that one can achieve impressive performance even in 2D tasks by explicitly modeling the 3D scene through a \emph{volumetric} field. We also take inspiration from Computer Graphics, where global, volumetric \emph{deformation fields} have been proposed as tools for manipulating implicitly represented objects \cite{sugihara2010warpcurves,seyb2019non}. Finally, we draw inspiration from photogrammetry, where reconstruction uncertainty is often modeled by placing Gaussian distributions on the spatial positions of identified feature points (see Fig.~\ref{fig:idea}).

\subsection{Intuition}

\begin{wrapfigure}[6]{r}{1.0in}
    \vspace{-15pt}
    \hspace{-15pt}
    \includegraphics{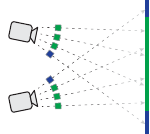}
    \hspace{-15pt}
\end{wrapfigure}
The intuition behind our re-parametrization is better seen with the simple scene shown in the inset. Consider a single solid blue segment with a green center embedded in the 2D plane. Imagine that this object is observed by two simplified cameras that capture rays in a 60-degree cone, and let us now consider the NeRF reconstruction problem as stated on this small dataset.

\begin{wrapfigure}[7]{l}{0.8in}
    \vspace{-10pt}
    \includegraphics{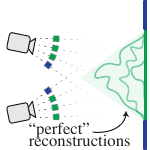}
    \hspace{-15pt}
\end{wrapfigure}
Trivially, the problem is an under-determined one: as shown in the inset, the green segment could be substituted by many possible curves while still resulting in a ``perfect'' photometric reconstruction according to the four pixels in our dataset. Indeed, there is a whole \emph{null-space} of solutions (green shaded region) to which the green segment could be perturbed without affecting the reconstruction loss\footnote{Note this is analogous to the condition number of near/far-field triangulation from photogrammetry that was discussed in \Cref{sec:intro}.}, and a NeRF trained on this dataset may converge to any one of these configurations depending on the training seed. Hence, one may quantify the uncertainty of a trained NeRF by asking ``how much can one perturb it without hurting the reconstruction accuracy?''

\begin{wrapfigure}[5]{r}{1.0in}
    \vspace{-10pt}
    \hspace{-15pt}
    \includegraphics{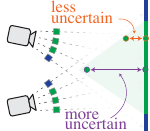}
\end{wrapfigure}
Crucially, this quantity will vary spatially: some regions of space will be more constrained by the training set and will allow only a limited perturbation before having an adverse effect on the loss (e.g., the edges of the segment, orange in the inset) while others will be much more uncertain (e.g., the middle of the segment, purple in the inset). Hence, we will be able to quantify the \emph{spatial} uncertainty of any trained NeRF by asking ``which \emph{regions} can one perturb without hurting the reconstruction accuracy?'' 

\begin{wrapfigure}[6]{l}{0.8in}
    \vspace{-15pt}
    \includegraphics{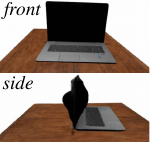}
    \hspace{-15pt}
\end{wrapfigure}
This quantity will be helpful beyond simple didactic examples: indeed, even general 3D scenes like the one in the inset can seem like pixel-perfect reconstructions from all training camera views (in this case, we trained a Nerfacto \cite{nerfstudio} model for 30,000 epochs with 40 front and back views of the laptop) but reveal large geometric artifacts when seen from a different angle.

\subsection{Modeling perturbations}
Inspired by all the considerations above, we introduce a deformation field $\deformfield:\mathbb{R}^D\rightarrow \mathbb{R}^D$, which one can interpret as a block that is ran on the input coordinates before the NeRF network. We choose a spatially meaningful parametrization in the form of vector displacements stored on the vertices of a grid of length $\gridSize$, allowing $\deformparam$ to be represented as a matrix $\deformparam\in\mathbb{R}^{\gridSize^D\times D}$, and defining a deformation for every spatial coordinate via trilinear interpolation
\begin{equation}
\deformfield_{\deformparam(\x)} = \text{Trilinear}(\x, \deformparam)\,.
\end{equation}

We can now reparametrize the radiance field $\rfield$ with $\theta$ by perturbing each coordinate $\x$ before applying the \emph{already-optimized} NeRF neural network
\begin{align}
\tilde\density_{\deformparam}(\x) &= \density_{\nerfparamsOptimal}(\x + \deformfield_{\deformparam}(\x))\,,\\ \tilde\radiance_{\deformparam}(\x) &= \radiance_{\nerfparamsOptimal}(\x + \deformfield_{\deformparam}(\x), \dir)\,,
\end{align}
resulting in the predicted pixel colors
\[
    \tilde\raycolor_{\deformparam}(\ray) = \sum_i \exp\left(-\sum_{j<i} \tilde\tau_j \delta_j\right)(1-\exp(-\tilde\tau_i\delta_i)) \: \tilde\radiance_i\, .
\]

We proceed by assuming a likelihood of the same form as with the NeRF parametrization, $\tilde\raycolor_{\deformparam}\given\deformparam\sim \mathcal{N}(\raycolor^\gt_n,\tfrac{1}{2})$. Under our assumption that $\nerfparamsOptimal$ are the optimal parameters obtained from NeRF training, it would be unreasonable to expect any non-trivial deformation to decrease the reconstruction loss; thus, it makes sense to place a regularizing independent Gaussian prior $\deformparam\sim \mathcal{N}(\mathbf{0},\lambda^{-1})$ on our new parameters and formulating the posterior $p(\deformparam\given I)$  whose negative log-likelihood $h(\deformparam)$ is given by
\begin{equation}
    h(\deformparam) = \expect_n 
    \expect_{\ray \sim \image_n}
\| \tilde\raycolor_{\deformparam}(\ray) - \raycolor^\gt_n(\ray) \|_2^2 
+ \lambda\|\deformparam\|^2\,.
\label{eq:nlltheta}\end{equation}

The minimum of \eq{nlltheta} must be obtained when $\deformparam=0$, as in that case $\tilde\density_{0}(\x) = \density_{\nerfparamsOptimal}(\x)$, $\tilde\radiance_{0}(\x) = \radiance_{\nerfparamsOptimal}(\x, \dir)$ and thus $\tilde\raycolor_{0}(\ray) = \raycolor_{\nerfparams}(\ray)$. Thus, zero is the mode of the distribution $p(\theta\given I)$ and we are finally in the ideal conditions for a Laplace approximation around $\deformparamOptimal=0$. Following Sec.~\ref{sec:lap}, this result in a distribution $\theta\sim\mathcal{N}(\mathbf{0},\covariance)$ where
\begin{equation}
    \covariance = - \Hessian(\mathbf{0})^{-1}
\end{equation}
where $\Hessian$ is the Hessian matrix of second derivatives of $h(\deformparam)$ evaluated at zero. Computing these second derivatives is a computationally intensive task; however, as we show below, a combination of statistical and NeRF-specific tools allows us to approximate it in terms of first derivatives only.

\begin{figure}
\vspace{-10pt}
    \includegraphics[width=\linewidth]{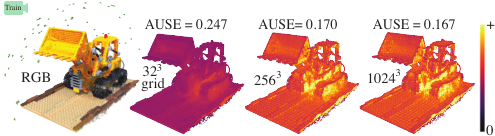}
    \vspace{-0.75cm}
    \caption{Very low resolutions may cause uncertainties to be underestimated, with diminishing returns for $\gridSize>256$.}
    \label{fig:ablation2}
    \vspace{-0.1cm}
\end{figure}

\subsection{Approximating \texorpdfstring{$\hessian$}{H}} %
For any parametric family of statistical distributions $p_{\params}$, the Hessian of the log-likelihood with respect to the parameters is also known as the \emph{Fisher information}
\begin{equation}
\Fisher(\params) = - \expected_{\mathbf{X}\sim p_\theta}\left[\frac{\partial^2 h(\mathbf{X}\,;\,\params)}{\partial\params^2}\middle|\params \right]   = - \hessian(\params)\,,
\label{eq:fisher}
\end{equation}
which (under reasonable regularity assumptions) can also be defined as the variance of the parametric score \cite[5.3]{LehmCase98}
\begin{equation}
    \Fisher(\params) = \expected_{\mathbf{X}\sim p_\theta}\left[\frac{\partial h(\mathbf{X}\,;\,\params)}{\partial\params}^\top\frac{\partial h(\mathbf{X}\,;\,\params)}{\partial\params} \middle| \params \right]\, .
\label{eq:fisher2}\end{equation}
Let us now denote the pair of random variables corresponding to a ray and its predicted color as $(\ray, \mathbf{y})$, where $\ray \sim \{\image_n\}$ and  $\gtcolor=\raycolor^\gt_n(\ray)$.
In our case, \eq{fisher2} takes the form
\begin{equation}
\Fisher(\params)
= \expect_{(\ray, \gtcolor)}
\left[
4 \: \residual_{\deformparam}(\ray) \:
\J_{\deformparam}(\ray)^\top \J_{\deformparam}(\ray) 
\right] + 2\lambda\identity
\end{equation}
where $\residual_{\deformparam}(\ray)$ is the ray residual error 
\[\residual_{\deformparam}(\ray) = \| \tilde\raycolor_{\deformparam}(\ray) - \raycolor^\gt_n(\ray) \|^2\] 
and $\J_{\deformparam}(\ray)$ is the Jacobian of first derivatives
\begin{equation}
    \J_{\deformparam}(\ray)=\frac{\partial \tilde\raycolor_{\deformparam}(\ray)}{\partial \deformparam}
\end{equation}
which can easily be computed via backpropagation. 

Further, as we typically do not have multiple observations of ray color for a single ray $\ray$, we can further simplify the above using the definition of conditional expectation
\begin{equation}
\Fisher(\params)
= \expect_{\ray}
\left[
4 \:
\expect_{\gtcolor \given \ray}
\left[
\residual_{\deformparam}(\ray)
\right]
\J_{\deformparam}(\ray)^\top \J_{\deformparam}(\ray) 
\right] + 2\lambda\identity\, ,
\end{equation}
noting that $\expect_{\gtcolor \given \ray}\left[ \residual_{\deformparam}(\ray) \right]$ is nothing more than $\tfrac{1}{2}$, the variance of our stated likelihood $\mathcal{N}(\raycolor^\gt_n,\tfrac{1}{2})$,
\begin{equation}
\Fisher(\params)
= \expect_{\ray}
\left[
2 \:
\J_{\deformparam}(\ray)^\top \J_{\deformparam}(\ray) 
\right] 
+ 2\lambda\identity\, .
\label{eq:variancedef2}
\end{equation}
Combining \eq{variancedef2} with \eq{fisher} and approximating the expectation via sampling of $R$ rays, we have our final expression for $\hessian$:
\begin{align}
\hessian(\theta) \approx -\tfrac{2}{\nRays}\sum_\ray \J_{\deformparam}(\ray)^\top \J_{\deformparam}(\ray) - 2 \lambda \identity  \,,
\label{eq:happrox}
\end{align}
It is worth remarking that, while $\hessian$ contains in it all the information that we will need to quantify the epistemic uncertainty of the given radiance field, its computation in \eq{happrox} \textit{does not} explicitly rely on the data from the training images but \textit{only} on information contained in the pre-trained model and the training camera parameters.

\subsection{Spatial uncertainty}
\label{subsec:spatial-uncertainty}

\begin{wrapfigure}[9]{r}{0.8in}
    \vspace{-45pt}
    \hspace{-15pt}
    \includegraphics{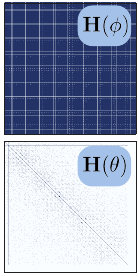}
\end{wrapfigure}
We can now fully take advantage of our proposed reparametrization. First, since each vector entry in $\deformparam$ corresponds to a vertex on our grid, its effect will be spatially limited to the cells containing it, making $\hessian (\deformparam)$ necessarily sparse and minimizing the number of correlated parameters (see inset, which compares the sparsity of $\hessian (\deformparam)$ to the that of an NeRF's  MLP parameters $\nerfparams$). In fact, thanks to this low number of correlations, we will proceed like \citet{Ritter2018ASL} and approximate $\covariance$ only through the diagonal entries of $\hessian$:
\begin{equation}
    \covariance \approx \text{diag}\left( \tfrac{2}{\nRays}\sum_\ray \J_{\deformparam}(\ray)^\top \J_{\deformparam}(\ray) + 2 \lambda \identity\right)^{-1}\, .
    \label{eq:final_hessian}
\end{equation}
Secondly, by measuring the variance of our deformation field~(intuitively, how much one could change the NeRF geometry without harming reconstruction quality), $\covariance$ critically encodes the \emph{spatial uncertainty} of the radiance field.
We can formalize this by considering the (root) diagonal entries of~$\covariance$, which define a marginal variance vector~$\boldsymbol{\sigma} = (\sigma_x,\sigma_y,\sigma_z)$. \silvianew{Much like in the photogrammetry works discussed in \Section{related} and \Figure{idea}, at each grid vertex, $\boldsymbol{\sigma}$ defines a spatial ellipsoid within which it can be deformed to minimal reconstruction cost.}
The norm of this vector $\sigma = \|\boldsymbol{\sigma}\|_2$ is then a positive scalar that measures the local spatial uncertainty of the radiance field at each grid vertex. Through it, we can define our \emph{spatial uncertainty field} $\ufield:\mathbb{R}^3\rightarrow\mathbb{R}^+$ given by
\begin{equation}
    \ufield(\x) = \text{Trilinear}(\x,\sigma)\,,
\end{equation}
\silvianew{which intuitively encodes how much the positioning of geometric region in our reconstruction can be trusted.} 
Strictly speaking, as defined above, $\mathcal{U}$ measures the uncertainty at $(1+\deformfield_{\deformparam})^{-1}(\x)$, not $\x$; however, we note that these are equivalent for the trained NeRF for which $\deformfield_{\deformparamOptimal}=0$.

The uncertainty field $\ufield$ is the main output of our algorithm and serves to illustrate the success of our approach. It is a first-of-its-kind theoretically derivated spatial measure of uncertainty that can be computed on \textit{any} NeRF architecture, without the need for additional training, expensive sampling or even access to the training images.
We will now validate it experimentally and show potential applications.
\vspace{-5pt}
\section{Experiments \& Applications}
\label{sec:results}

\begin{figure*}
\includegraphics[width=\linewidth]{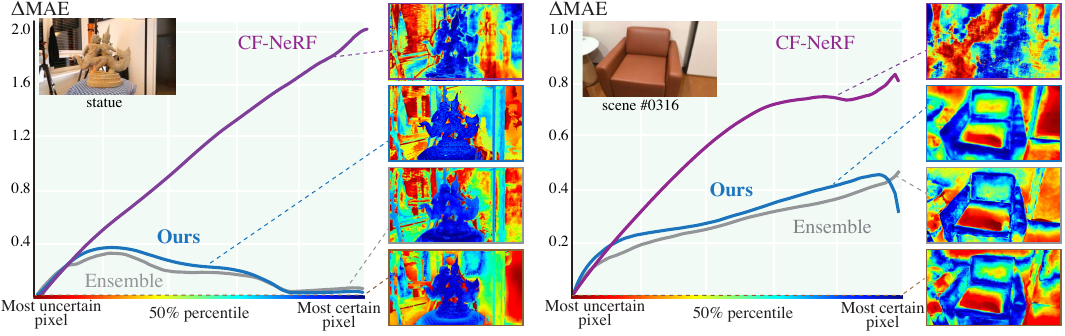}
\\[0.5em]
\centering
\resizebox{\linewidth}{!}{
\setlength{\tabcolsep}{13pt} %
\begin{tabular}{c|c|c|c|c|c|c|c|c|c|c|c}
\backslashbox{Method}{Scene} & \multicolumn{2}{c|}{africa} & \multicolumn{2}{c|}{basket} & \multicolumn{2}{c|}{statue} & torch & \#0000 &  \#0079 & \#0158 & \#0316 \\ \hline
Ensemble & \multicolumn{2}{c|}{0.18} & \multicolumn{2}{c|}{0.24} & \multicolumn{2}{c|}{0.15} & 0.19 & 0.28 & 0.36 & 0.19 & 0.26 \\ \hline \hline
CF-NeRF & \multicolumn{2}{c|}{0.35} & \multicolumn{2}{c|}{0.31} & \multicolumn{2}{c|}{0.46} & 0.97 & 0.59 & 0.43 & 0.55 & 0.54 \\ \hline
Ours & \multicolumn{2}{c|}{\textbf{0.27}} & \multicolumn{2}{c|}{\textbf{0.28}} & \multicolumn{2}{c|}{\textbf{0.17}} & \textbf{0.22} & \textbf{0.28} & \textbf{0.35} & \textbf{0.20} & \textbf{0.29} \\ \hline
\end{tabular}
}

\vspace{-0.2cm}
\caption{
    The uncertainties computed with our algorithm on the ScanNet and Light Field dataset are significantly more calibrated to the real NeRF depth error than the previous state-of-the-art CF-NeRF \cite{cfnerf}, even matching the performance of extremely costly ensembles. Images are colored by uncertainty / depth error \emph{percentile} instead of value to be comparable.
} %
\label{fig:depth-uncertainty}
\end{figure*}

We validate our theoretically-derived algorithm through our uncertainty field's correlation with the depth prediction error in a given NeRF~(\Section{exp1}), show a prototypical application to a NeRF clean-up task~(\Section{exp2}) and justify our parametric choices through ablation studies~(\Section{exp3}).

\textbf{Implementation.} 
Unless specified otherwise, all NeRFs used throughout this paper use Nerfstudio's Nerfacto \cite{nerfstudio} as the base architecture and are pre-trained for 30,000 steps. We extract our uncertainty field $\ufield$ using 1,000 random batches of 4,096 rays each sampled from a scene's training cameras, with $M=256$ and $\lambda= 10^{-4} / M^3$, in a process that takes around $90$ seconds on an NVIDIA RTX 6000.
Once computed for a given scene, our derived uncertainty field conveniently functions as an additional color channel that can be volumetrically rendered in a form (and cost) analogous to the usual RGB. For visualization clarity, all our results use a logarithmic scale, rendering $\log\ufield$ instead of $\ufield$.

\vspace{-0.3cm}
\subsection{Uncertainty Evaluation -- Figures~\ref{fig:depth-uncertainty}}
\label{sec:exp1}

We evaluate the estimated uncertainty of \algoName by showing its correlation with the NeRF depth error. We choose the error in predicted depth as the best signal that conveys NeRF's unreliability in geometric prediction, as RGB error has been shown to not be representative of true uncertainty due to radiance accumulation and model biases \cite{sunderhauf2022densityaware}. 

\paragraph{Metric} We measure correlation through the Area Under Sparsification Error (AUSE) \cite{ilg2018uncertainty, Bae2021}. The pixels in each test image are removed gradually (``sparsified'') twice: first, according to their respective depth error; second, by their uncertainty measure. The difference between the Mean Absolute depth Error ($\Delta$MAE) of the remaining pixels in both processes, at each stage, provides the sparsification curves. 

\paragraph{Data} In \Figure{depth-uncertainty}, we use 4 ScanNet scenes (\#0000-001, \#0079-000, \#0158-000, \#0316-000) with groundtruth depths provided. Each scene employs 40 images split into 5 test and 35 train images, following NerfingMVS \cite{wei2021nerfingmvs}. Additionally, we use 4 scenes from the Light Field dataset ~\cite{Ycer2016Efficient3D, zhang2020nerf} (torch, statue, basket, africa), with the same train-test split and pseudo-ground-truth depth map approach as CF-NeRF \cite{cfnerf}.

\paragraph{Baselines} For Figure \ref{fig:depth-uncertainty}, we display sparsification curves derived from our uncertainty field, with the previous state-of-the-art CF-NeRF \cite{cfnerf} and with the standard deviations obtained by the costly process of training an ensemble of ten identical yet differently seeded NeRFs.
Next to each graph, we visualize the depth error together with the~(ascending) per-pixels rank produced by each method~(i.e., the ordering that produces the curves). It is worth noting that, unlike CF-NeRF \cite{cfnerf}, we do not measure disparity error due to its heightened sensitivity to low-range depth errors and significant underestimation of errors in distant points.

\paragraph{Results} The results are consistent across Figure~\ref{fig:depth-uncertainty}. \algoName's uncertainty shows significant improvement in correlation with depth error compared to CF-NeRF~\cite{cfnerf}, both quantitatively and qualitatively. Further, our uncertainty is extremely close to the standard deviation of a costly ensemble in both AUSE and sparsification plots, while requiring no additional trained NeRFs, saving time and memory.

\subsection{NeRF Clean Up -- Figures \ref{fig:teaser} and \ref{fig:nerfbuster-comparison}}
\label{sec:exp2}

A common reconstruction artifact in NeRFs are ``Floaters'', often caused by a lack of information in training data.
These inherently correspond to regions of high uncertainty; therefore, we propose removing them by thresholding the scene according to our uncertainty field $\ufield$ during rendering.

In \Figure{nerfbuster-comparison}, we compare our algorithm's performance to the current state of the art for post-hoc floater removal, Nerfbusters \cite{warburg2023nerfbusters}, which employs a 3D diffusion model and a ``visibility mask'' to guide additional training steps during which some floaters are removed. For our comparison, we use the same dataset proposed by Nerfbusters along with their proposed metric of \emph{Coverage}, together with more common measures of image quality. An ideal clean-up would boost image quality while keeping pixel coverage high.

When using fixed threshold values like $0.9$ or $0.4$, \algoName obtains similar PSNR values to Nerfbusters while allowing for a higher coverage. If one selects the best possible threshold value for each scene out of ten equally spaced ones, \algoName outperforms Nerfbusters in both PSNR and coverage.
It is worth noting that \algoName achieves with a significantly lower computational footprint: unlike Nerfbusters, we do not require storing and evaluating a 3D diffusion model, we are faster (96 seconds vs 20 minutes) by eliminating the need for additional training and we circumvent the use of a ``visibility mask'' altogether by storing all necessary information in our computed Hessian $\hessian$.

Our qualitative results in ~\Figure{nerfbuster-comparison} show that our method can filter floaters that are missed by Nerfbusters (\textit{Century}), is not prone to sampling artifacts caused by floater removal~(\textit{Flowers}) and provides the parametric flexibility necessary to avoid over-filtering~(\textit{Pikachu}).

\begin{figure}
    \includegraphics[width=\linewidth]{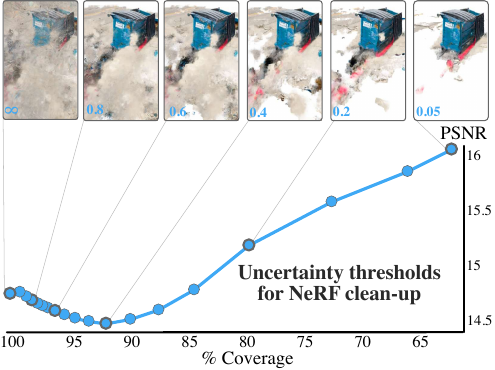}
    \vspace{-0.6cm}
    \caption{Applying different thresholds to uncertainty for the NeRF clean up task on the ``garbage'' scene from Nerfbusters dataset~\cite{warburg2023nerfbusters}. Rightmost image (threshold=$\infty$) shows the original.}
    \label{fig:floater-threshold}
\end{figure}

\subsection{Algorithmic ablations -- Figures \ref{fig:ablation1}, \ref{fig:ablation2} and \ref{fig:floater-threshold}}
\label{sec:exp3}

In \Section{method}, we justified our introduction of the perturbation field $\deformfield$ partly through the desire to make our algorithm independent to the specific NeRF architecture used. In \Figure{ablation1}, we show that this is indeed the case, as we obtain qualitatively similar results for three representatively different architectures (Mip NeRF \cite{mipnerf}, Instant NGP \cite{instantngp} and Nerfacto \cite{nerfstudio}) on the ``Lego'' scene from the NeRF Synthetic dataset \cite{mildenhall2020nerf} with 60 training views from its left hemisphere. 

This success introduces an algorithmic choice; namely, the discretization of the deformation field. In \Section{method}, we propose storing it in a uniform spatial grid of size $\gridSize^3$ from which deformation values can be trilinearly interpolated. The value of $\gridSize$ thus becomes an algorithmic parameter, which we explore for the same example in \Figure{ablation2}. We find that surface uncertainty can be missed for small values of~$\gridSize$, resulting in a generally more certain map that is only activated on points of very high uncertainty, with diminishing returns being obtained for larger, more costly $M>256$.

Finally, our algorithm's flagship application to NeRF artifact removal also contains a parameter choice, in the form of the uncertainty threshold. As we show in \Figure{floater-threshold}, decreasing this parameter can a gradually clean a floater-heavy scene leaving a floater-free clean capture of the target object. Since our uncertainty field only needs to be computed once, we suggest that this threshold can serve as \textit{real-time} user control in interactive NeRF setups like Nerfstudio \cite{nerfstudio}.

\section{Conclusions}

\begin{figure}
    \includegraphics[width=\linewidth]{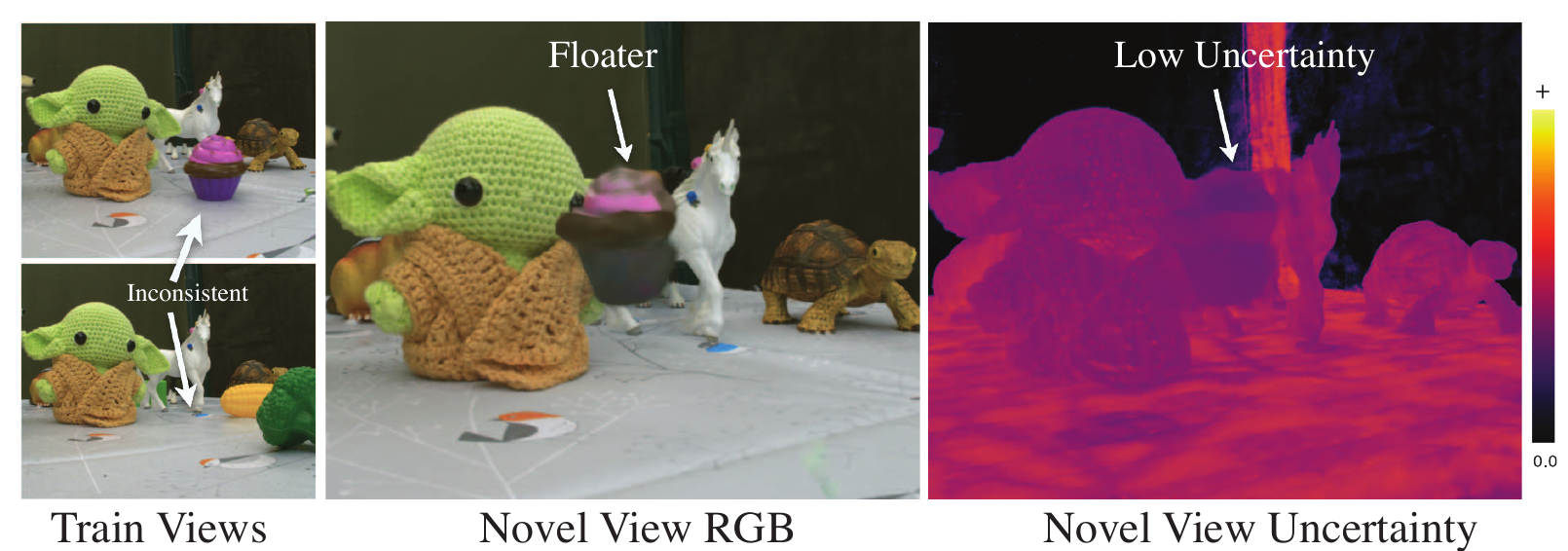}
    \vspace{-0.6cm}
    \caption{\algoName quantifies only epistemic uncertainty in NeRF and is thus unable to capture aleatoric effects like those stemming from training inconsistencies.}
    \label{fig:limitation}
\end{figure}

We have introduced \algoName, an algorithm to quantify the uncertainty of any trained Neural Radiance Field without independently of its architecture and without additional training nor access to the original training images. Our algorithm outputs a spatial uncertainty field, which we have shown is meaningfully correlated to the NeRF depth error and can be thresholded for use in applications like NeRF cleanup.

We discretize our spatial deformation field using a uniform grid, which can lead to a high memory cost being incurred in regions of little geometric interest. Future work may significantly improve our performance by considering more complex hierarchical data structures like octrees.
Separately, in our algorithm's derivation, we focus only on the diagonal of $\hessian$, disregarding (minimal) inter-parametric correlations. Future applications may require their inclusion, possibly through low-rank matrix decompositions \cite{ambikasaran2013mathcal,ambikasaran2014fast}.

At a higher level, our algorithm's stated goal is to capture only the epistemic uncertainty of the NeRF, often present through missing or occluded data. As such, aleatoric uncertainty caused by noise or inconsistencies between views is not captured by our method (see \Figure{limitation}).
We are optimistic that combining our work with current frameworks for aleatoric quantification like \cite{robustnerf, nerfwild} will result in a complete study of all sources of uncertainty in NeRF.

More broadly, our algorithm is limited to quantifying the uncertainty of NeRFs, and cannot be trivially translated to other frameworks. Nonetheless, we look forward to similar deformation-based Laplace approximations being formulated for more recent spatial representations like 3D Gaussian splatting~\cite{kerbl20233d}. 
As the Deep Learning revolution takes Computer Vision algorithms to new horizons of performance and increasingly critical applications, we hope that works like ours can aid in understanding what our models know and do not know as well as the confidence of their guesses.

\section*{Acknowledgements}
We would like to thank Alireza Mousavi-Hosseini, Agustinus Kristiadi, Towaki Takikawa, Kevin Swersky, Richard Szeliski, Aaron Hertzmann, Georgios Kopanas, Vladimir Kim, and Nathan Carr for helpful discussions and feedback.
The third author is funded in part by an NSERC Vanier Scholarship and an Adobe Research Fellowship.

{
    \small
    \bibliographystyle{ieeenat_fullname}
    \bibliography{main}

\begin{thebibliography}{59}
\providecommand{\natexlab}[1]{#1}
\providecommand{\url}[1]{\texttt{#1}}
\expandafter\ifx\csname urlstyle\endcsname\relax
  \providecommand{\doi}[1]{doi: #1}\else
  \providecommand{\doi}{doi: \begingroup \urlstyle{rm}\Url}\fi

\bibitem[Abdar et~al.(2021)Abdar, Pourpanah, Hussain, Rezazadegan, Liu,
  Ghavamzadeh, Fieguth, Khosravi, Acharya, Makarenkov, and
  Nahavandi]{uncertaintyReview2021}
Moloud Abdar, Farhad Pourpanah, Sadiq Hussain, Dana Rezazadegan, Li Liu,
  Mohammad Ghavamzadeh, Paul Fieguth, Abbas Khosravi, U~Rajendra Acharya,
  Vladimir Makarenkov, and Saeid Nahavandi.
\newblock A review of uncertainty quantification in deep learning: Techniques,
  applications and challenges.
\newblock \emph{arXive preprint at arXiv:2011.06225}, 2021.

\bibitem[Ambikasaran and Darve(2013)]{ambikasaran2013mathcal}
Sivaram Ambikasaran and Eric Darve.
\newblock An$\backslash$mathcal o (n$\backslash$log n) o (n log n) fast direct
  solver for partial hierarchically semi-separable matrices: With application
  to radial basis function interpolation.
\newblock \emph{Journal of Scientific Computing}, 2013.

\bibitem[Ambikasaran et~al.(2014)Ambikasaran, O'Neil, and
  Singh]{ambikasaran2014fast}
Sivaram Ambikasaran, Michael O'Neil, and Karan~Raj Singh.
\newblock Fast symmetric factorization of hierarchical matrices with
  applications.
\newblock \emph{arXiv preprint arXiv:1405.0223}, 2014.

\bibitem[Ayhan and Berens(2022)]{ayhan2022test}
Murat~Seckin Ayhan and Philipp Berens.
\newblock Test-time data augmentation for estimation of heteroscedastic
  aleatoric uncertainty in deep neural networks.
\newblock \emph{Medical Imaging with Deep Learning}, 2022.

\bibitem[Bae et~al.(2021)Bae, Budvytis, and Cipolla]{Bae2021}
Gwangbin Bae, Ignas Budvytis, and Roberto Cipolla.
\newblock Estimating and exploiting the aleatoric uncertainty in surface normal
  estimation.
\newblock \emph{ICCV}, 2021.

\bibitem[Barron et~al.(2021{\natexlab{a}})Barron, Mildenhall, Tancik, Hedman,
  Martin-Brualla, and Srinivasan]{mipnerf}
Jonathan~T. Barron, Ben Mildenhall, Matthew Tancik, Peter Hedman, Ricardo
  Martin-Brualla, and Pratul~P. Srinivasan.
\newblock Mip-nerf: A multiscale representation for anti-aliasing neural
  radiance fields.
\newblock \emph{ICCV}, 2021{\natexlab{a}}.

\bibitem[Barron et~al.(2021{\natexlab{b}})Barron, Mildenhall, Verbin,
  Srinivasan, and Hedman]{mipnerf-360}
Jonathan~T. Barron, Ben Mildenhall, Dor Verbin, Pratul~P. Srinivasan, and Peter
  Hedman.
\newblock Mip-nerf 360: Unbounded anti-aliased neural radiance fields.
\newblock \emph{CVPR}, 2021{\natexlab{b}}.

\bibitem[Bartoli(2003)]{Bartoli2003}
Adrien Bartoli.
\newblock Towards gauge invariant bundle adjustment: A solution based on gauge
  dependent damping.
\newblock \emph{ICCV}, 2003.

\bibitem[Blundell et~al.(2015)Blundell, Cornebise, Kavukcuoglu, and
  Wierstra]{blundell2015weight}
Charles Blundell, Julien Cornebise, Koray Kavukcuoglu, and Daan Wierstra.
\newblock Weight uncertainty in neural networks.
\newblock \emph{ICML}, 2015.

\bibitem[Bojanowski et~al.(2019)Bojanowski, Joulin, Lopez-Paz, and
  Szlam]{bojanowski2019optimizing}
Piotr Bojanowski, Armand Joulin, David Lopez-Paz, and Arthur Szlam.
\newblock Optimizing the latent space of generative networks.
\newblock \emph{ICML}, 2019.

\bibitem[Chitta et~al.(2021)Chitta, Prakash, and Geiger]{chitta2021neat}
Kashyap Chitta, Aditya Prakash, and Andreas Geiger.
\newblock Neat: Neural attention fields for end-to-end autonomous driving.
\newblock \emph{ICCV}, 2021.

\bibitem[Daxberger et~al.(2021)Daxberger, Kristiadi, Immer, Eschenhagen, Bauer,
  and Hennig]{Kristiadi2021}
Erik Daxberger, Agustinus Kristiadi, Alexander Immer, Runa Eschenhagen,
  Matthias Bauer, and Philipp Hennig.
\newblock Laplace redux -- effortless bayesian deep learning.
\newblock \emph{NeuRIPS}, 2021.

\bibitem[Ellis et~al.(1980)Ellis, Gaillard, Nanopoulos, and
  Rudaz]{ellis1980uncertainties}
John Ellis, Mary~K Gaillard, Dimitri~V Nanopoulos, and Serge Rudaz.
\newblock Uncertainties in the proton lifetime.
\newblock \emph{Nuclear Physics B}, 1980.

\bibitem[Hoffman et~al.(2022)Hoffman, Le, Sountsov, Suter, Lee, Mansinghka, and
  Saurous]{probnerf}
Matthew~D. Hoffman, Tuan~Anh Le, Pavel Sountsov, Christopher Suter, Ben Lee,
  Vikash~K. Mansinghka, and Rif~A. Saurous.
\newblock Probnerf: Uncertainty-aware inference of 3d shapes from 2d images.
\newblock \emph{AISTATS}, 2022.

\bibitem[Ilg et~al.(2018)Ilg, Özgün Çiçek, Galesso, Klein, Makansi, Hutter,
  and Brox]{ilg2018uncertainty}
Eddy Ilg, Özgün Çiçek, Silvio Galesso, Aaron Klein, Osama Makansi, Frank
  Hutter, and Thomas Brox.
\newblock Uncertainty estimates and multi-hypotheses networks for optical flow.
\newblock \emph{ECCV}, 2018.

\bibitem[Jin et~al.(2023)Jin, Chen, Rückin, and Popovic]{Jin2023neunbv}
Liren Jin, Xieyuanli Chen, Julius Rückin, and Marija Popovic.
\newblock Neu-nbv: Next best view planning using uncertainty estimation in
  image-based neural rendering.
\newblock \emph{IROS}, 2023.

\bibitem[Kendall and Gal(2017)]{kendall2017uncertainties}
Alex Kendall and Yarin Gal.
\newblock What uncertainties do we need in bayesian deep learning for computer
  vision?
\newblock \emph{NeuRIPS}, 2017.

\bibitem[Kerbl et~al.(2023)Kerbl, Kopanas, Leimk{\"u}hler, and
  Drettakis]{kerbl20233d}
Bernhard Kerbl, Georgios Kopanas, Thomas Leimk{\"u}hler, and George Drettakis.
\newblock 3d gaussian splatting for real-time radiance field rendering.
\newblock \emph{ACM Trans. Graph.}, 2023.

\bibitem[Lakshminarayanan et~al.(2016)Lakshminarayanan, Pritzel, and
  Blundell]{ensemble2016}
Balaji Lakshminarayanan, Alexander Pritzel, and Charles Blundell.
\newblock Simple and scalable predictive uncertainty estimation using deep
  ensembles.
\newblock \emph{NeuRIPS}, 2016.

\bibitem[Lehmann and Casella(1998)]{LehmCase98}
Erich~L. Lehmann and George Casella.
\newblock \emph{Theory of Point Estimation}.
\newblock Springer-Verlag, second edition, 1998.

\bibitem[Marimont and Tarroni(2021)]{marimont2021implicit}
Sergio~Naval Marimont and Giacomo Tarroni.
\newblock Implicit field learning for unsupervised anomaly detection in medical
  images.
\newblock \emph{MICCAI}, 2021.

\bibitem[Martin-Brualla et~al.(2020)Martin-Brualla, Radwan, Sajjadi, Barron,
  Dosovitskiy, and Duckworth]{nerfwild}
Ricardo Martin-Brualla, Noha Radwan, Mehdi Sajjadi, Jonathan~T. Barron, Alexey
  Dosovitskiy, and Daniel Duckworth.
\newblock {NeRF in the Wild}: Neural radiance fields for unconstrained photo
  collections.
\newblock \emph{CVPR}, 2020.

\bibitem[Mildenhall et~al.(2020{\natexlab{a}})Mildenhall, Srinivasan, Tancik,
  Barron, Ramamoorthi, and Ng]{NeRF}
Ben Mildenhall, Pratul~P. Srinivasan, Matthew Tancik, Jonathan~T. Barron, Ravi
  Ramamoorthi, and Ren Ng.
\newblock {NeRF}: Representing scenes as neural radiance fields for view
  synthesis.
\newblock \emph{ECCV}, 2020{\natexlab{a}}.

\bibitem[Mildenhall et~al.(2020{\natexlab{b}})Mildenhall, Srinivasan, Tancik,
  Barron, Ramamoorthi, and Ng]{mildenhall2020nerf}
Ben Mildenhall, Pratul~P. Srinivasan, Matthew Tancik, Jonathan~T. Barron, Ravi
  Ramamoorthi, and Ren Ng.
\newblock Nerf: Representing scenes as neural radiance fields for view
  synthesis.
\newblock \emph{ECCV}, 2020{\natexlab{b}}.

\bibitem[Monteiro et~al.(2020)Monteiro, Le~Folgoc, Coelho~de Castro, Pawlowski,
  Marques, Kamnitsas, van~der Wilk, and Glocker]{monteiro2020stochastic}
Miguel Monteiro, Lo{\"\i}c Le~Folgoc, Daniel Coelho~de Castro, Nick Pawlowski,
  Bernardo Marques, Konstantinos Kamnitsas, Mark van~der Wilk, and Ben Glocker.
\newblock Stochastic segmentation networks: Modelling spatially correlated
  aleatoric uncertainty.
\newblock \emph{Advances in neural information processing systems}, 2020.

\bibitem[Morris et~al.(1999)Morris, Kanatani, and
  Kanade]{Morris1999UncertaintyMF}
Daniel Morris, Kenichi Kanatani, and Takeo Kanade.
\newblock Uncertainty modeling for optimal structure from motion.
\newblock \emph{Workshop on Vision Algorithms}, 1999.

\bibitem[M\"uller et~al.(2022)M\"uller, Evans, Schied, and Keller]{instantngp}
Thomas M\"uller, Alex Evans, Christoph Schied, and Alexander Keller.
\newblock Instant neural graphics primitives with a multiresolution hash
  encoding.
\newblock \emph{SIGGRAPH}, 2022.

\bibitem[Neal(1995)]{Neal1995BayesianLF}
Radford~M. Neal.
\newblock \emph{Bayesian Learning for Neural Networks}.
\newblock Springer-Verlag, 1995.

\bibitem[Niemeyer et~al.(2021)Niemeyer, Barron, Mildenhall, Sajjadi, Geiger,
  and Radwan]{niemeyer2021regnerf}
Michael Niemeyer, Jonathan~T. Barron, Ben Mildenhall, Mehdi S.~M. Sajjadi,
  Andreas Geiger, and Noha Radwan.
\newblock Regnerf: Regularizing neural radiance fields for view synthesis from
  sparse inputs.
\newblock \emph{CVPR}, 2021.

\bibitem[Palmer(2000)]{palmer2000predicting}
Tim~N Palmer.
\newblock Predicting uncertainty in forecasts of weather and climate.
\newblock \emph{Reports on progress in Physics}, 2000.

\bibitem[Pan et~al.(2022)Pan, Lai, Song, and Huang]{pan2022activenerf}
Xuran Pan, Zihang Lai, Shiji Song, and Gao Huang.
\newblock Activenerf: Learning where to see with uncertainty estimation.
\newblock \emph{ECCV}, 2022.

\bibitem[Park et~al.(2020)Park, Sinha, Barron, Bouaziz, Goldman, Seitz, and
  Martin{-}Brualla]{nerfies}
Keunhong Park, Utkarsh Sinha, Jonathan~T. Barron, Sofien Bouaziz, Dan~B.
  Goldman, Steven~M. Seitz, and Ricardo Martin{-}Brualla.
\newblock Deformable neural radiance fields.
\newblock \emph{ICCV}, 2020.

\bibitem[Park et~al.(2021)Park, Sinha, Hedman, Barron, Bouaziz, Goldman,
  Martin-Brualla, and Seitz]{park2021hypernerf}
Keunhong Park, Utkarsh Sinha, Peter Hedman, Jonathan~T. Barron, Sofien Bouaziz,
  Dan~B Goldman, Ricardo Martin-Brualla, and Steven~M. Seitz.
\newblock Hypernerf: A higher-dimensional representation for topologically
  varying neural radiance fields.
\newblock \emph{ACM Trans. Graph.}, 2021.

\bibitem[Park et~al.(2023)Park, Henzler, Mildenhall, Barron, and
  Martin-Brualla]{park2023camp}
Keunhong Park, Philipp Henzler, Ben Mildenhall, Jonathan~T. Barron, and Ricardo
  Martin-Brualla.
\newblock Camp: Camera preconditioning for neural radiance fields, 2023.

\bibitem[Ran et~al.(2023)Ran, Zeng, He, Chen, Li, Chen, Lee, and Ye]{Ran_2023}
Yunlong Ran, Jing Zeng, Shibo He, Jiming Chen, Lincheng Li, Yingfeng Chen,
  Gimhee Lee, and Qi Ye.
\newblock {NeurAR}: Neural uncertainty for autonomous 3d reconstruction with
  implicit neural representations.
\newblock \emph{{IEEE} Robotics and Automation Letters}, 2023.

\bibitem[Rebain et~al.(2022)Rebain, Matthews, Yi, Lagun, and
  Tagliasacchi]{rebain2022lolnerf}
Daniel Rebain, Mark Matthews, Kwang~Moo Yi, Dmitry Lagun, and Andrea
  Tagliasacchi.
\newblock Lolnerf: Learn from one look.
\newblock \emph{CVPR}, 2022.

\bibitem[Ritter et~al.(2018)Ritter, Botev, and Barber]{Ritter2018ASL}
Hippolyt Ritter, Aleksandar Botev, and David Barber.
\newblock A scalable laplace approximation for neural networks.
\newblock \emph{ICLR}, 2018.

\bibitem[Sabour et~al.(2023)Sabour, Vora, Duckworth, Krasin, Fleet, and
  Tagliasacchi]{robustnerf}
Sara Sabour, Suhani Vora, Daniel Duckworth, Ivan Krasin, David~J. Fleet, and
  Andrea Tagliasacchi.
\newblock Robustnerf: Ignoring distractors with robust losses.
\newblock \emph{CVPR}, 2023.

\bibitem[Schecher and Driscoll(1988)]{schecher1988evaluation}
William~D Schecher and Charles~T Driscoll.
\newblock An evaluation of the equilibrium calculations within acidification
  models: the effect of uncertainty in measured chemical components.
\newblock \emph{Water Resources Research}, 1988.

\bibitem[Seyb et~al.(2019)Seyb, Jacobson, Nowrouzezahrai, and
  Jarosz]{seyb2019non}
Dario Seyb, Alec Jacobson, Derek Nowrouzezahrai, and Wojciech Jarosz.
\newblock Non-linear sphere tracing for rendering deformed signed distance
  fields.
\newblock \emph{ACM Trans. Graph.}, 2019.

\bibitem[Shen et~al.(2021)Shen, Ruiz, Agudo, and Moreno-Noguer]{stochasticnerf}
Jianxiong Shen, Adria Ruiz, Antonio Agudo, and Francesc Moreno-Noguer.
\newblock Stochastic neural radiance fields: Quantifying uncertainty in
  implicit 3d representations.
\newblock \emph{3DV}, 2021.

\bibitem[Shen et~al.(2022)Shen, Agudo, Moreno-Noguer, and Ruiz]{cfnerf}
Jianxiong Shen, Antonio Agudo, Francesc Moreno-Noguer, and Adria Ruiz.
\newblock Conditional-flow nerf: Accurate 3d modelling with reliable
  uncertainty quantification.
\newblock \emph{ECCV}, 2022.

\bibitem[Smith(2013)]{smith2013uncertainty}
Ralph~C Smith.
\newblock \emph{Uncertainty quantification: theory, implementation, and
  applications}.
\newblock Siam, 2013.

\bibitem[Sugihara et~al.(2010)Sugihara, Wyvill, and
  Schmidt]{sugihara2010warpcurves}
Masamichi Sugihara, Brian Wyvill, and Ryan Schmidt.
\newblock Warpcurves: A tool for explicit manipulation of implicit surfaces.
\newblock \emph{Computers \& Graphics}, 2010.

\bibitem[Szeliski(2022)]{szeliski2022computer}
Richard Szeliski.
\newblock \emph{Computer vision: algorithms and applications}.
\newblock Springer Nature, 2022.

\bibitem[Szeliski and Kang(1997)]{Szeliski1997}
Richard Szeliski and Sing~Bing Kang.
\newblock Shape ambiguities in structure from motion.
\newblock \emph{Pattern Analysis and Machine Intelligence, IEEE Transactions
  on}, 1997.

\bibitem[Sünderhauf et~al.(2023)Sünderhauf, Abou-Chakra, and
  Miller]{sunderhauf2022densityaware}
Niko Sünderhauf, Jad Abou-Chakra, and Dimity Miller.
\newblock Density-aware nerf ensembles: Quantifying predictive uncertainty in
  neural radiance fields.
\newblock \emph{ICRA}, 2023.

\bibitem[Tagliasacchi and Mildenhall(2022)]{digest}
Andrea Tagliasacchi and Ben Mildenhall.
\newblock Volume rendering digest (for nerf).
\newblock \emph{arXiv preprint arXiv:2209.02417}, 2022.

\bibitem[Tancik et~al.(2023)Tancik, Weber, Ng, Li, Yi, Wang, Kristoffersen,
  Austin, Salahi, Ahuja, Mcallister, Kerr, and Kanazawa]{nerfstudio}
Matthew Tancik, Ethan Weber, Evonne Ng, Ruilong Li, Brent Yi, Terrance Wang,
  Alexander Kristoffersen, Jake Austin, Kamyar Salahi, Abhik Ahuja, David
  Mcallister, Justin Kerr, and Angjoo Kanazawa.
\newblock Nerfstudio: A modular framework for neural radiance field
  development.
\newblock \emph{Special Interest Group on Computer Graphics and Interactive
  Techniques Conference Conference Proceedings}, 2023.

\bibitem[Tkach et~al.(2017)Tkach, Tagliasacchi, Remelli, Pauly, and
  Fitzgibbon]{tkach2017online}
Anastasia Tkach, Andrea Tagliasacchi, Edoardo Remelli, Mark Pauly, and Andrew
  Fitzgibbon.
\newblock Online generative model personalization for hand tracking.
\newblock \emph{ACM Trans. Graph.}, 2017.

\bibitem[Triggs et~al.(1999)Triggs, McLauchlan, Hartley, and
  Fitzgibbon]{Triggs1999BundleA}
Bill Triggs, Philip~F. McLauchlan, Richard~I. Hartley, and Andrew~William
  Fitzgibbon.
\newblock Bundle adjustment - a modern synthesis.
\newblock \emph{Workshop on Vision Algorithms}, 1999.

\bibitem[Warburg et~al.(2023)Warburg, Weber, Tancik, Holynski, and
  Kanazawa]{warburg2023nerfbusters}
Frederik Warburg, Ethan Weber, Matthew Tancik, Aleksander Holynski, and Angjoo
  Kanazawa.
\newblock Nerfbusters: Removing ghostly artifacts from casually captured nerfs.
\newblock \emph{ICCV}, 2023.

\bibitem[Wei et~al.(2021)Wei, Liu, Rao, Zhao, Lu, and Zhou]{wei2021nerfingmvs}
Yi Wei, Shaohui Liu, Yongming Rao, Wang Zhao, Jiwen Lu, and Jie Zhou.
\newblock Nerfingmvs: Guided optimization of neural radiance fields for indoor
  multi-view stereo.
\newblock \emph{ICCV}, 2021.

\bibitem[Wilson and Wehrwein(2020)]{Wilson2020VisualizingSB}
Kyle Wilson and Scott Wehrwein.
\newblock Visualizing spectral bundle adjustment uncertainty.
\newblock \emph{2020 International Conference on 3D Vision (3DV)}, 2020.

\bibitem[Yan et~al.(2023)Yan, Yang, and Zha]{yan2023active}
Zike Yan, Haoxiang Yang, and Hongbin Zha.
\newblock Active neural mapping.
\newblock \emph{arXiv preprint arXiv:2308.16246}, 2023.

\bibitem[Y{\"u}cer et~al.(2016)Y{\"u}cer, Sorkine-Hornung, and
  Disney]{Ycer2016Efficient3D}
Kaan Y{\"u}cer, Alexander Sorkine-Hornung, and Oliver Disney.
\newblock Efficient 3 d object segmentation from densely sampled light fields
  with applications to 3 d reconstruction.
\newblock \emph{ACM TOG}, 2016.

\bibitem[Zaidi et~al.(2021)Zaidi, Zela, Elsken, Holmes, Hutter, and
  Teh]{ensemble2021}
Sheheryar Zaidi, Arber Zela, Thomas Elsken, Christopher~C. Holmes, Frank
  Hutter, and Yee~Whye Teh.
\newblock Neural ensemble search for uncertainty estimation and dataset shift.
\newblock \emph{NeuRIPS}, 2021.

\bibitem[Zhan et~al.(2022)Zhan, Zheng, Xu, Reid, and
  Rezatofighi]{zhan2022activermap}
Huangying Zhan, Jiyang Zheng, Yi Xu, Ian Reid, and Hamid Rezatofighi.
\newblock Activermap: Radiance field for active mapping and planning.
\newblock \emph{arXiv preprint arXiv:2211.12656}, 2022.

\bibitem[Zhang et~al.(2020)Zhang, Riegler, Snavely, and Koltun]{zhang2020nerf}
Kai Zhang, Gernot Riegler, Noah Snavely, and Vladlen Koltun.
\newblock Nerf++: Analyzing and improving neural radiance fields.
\newblock \emph{preprint arXiv:2010.07492}, 2020.

\end{thebibliography}
}

\end{document}